\documentclass[conference]{IEEEtran}
\makeatletter

\def\ps@IEEEtitlepagestyle{%
  \def\@oddfoot{\mycopyrightnotice}%
  \def\@evenfoot{}%
}
\def\mycopyrightnotice{%
    {\footnotesize 979-8-3315-3559-9/25/\$31.00~\copyright~2025 IEEE\hfill}
  \gdef\mycopyrightnotice{}
}

\usepackage{cite}
\usepackage{amsmath,amssymb,amsfonts}
\usepackage{algorithmic}
\usepackage{graphicx}
\usepackage{pgfplots}

\usepackage{textcomp}
\usepackage{xcolor}
\usepackage{adjustbox}

\usepackage{physics}
\usepackage{tikz}
\usepackage{tikz-3dplot}

\usetikzlibrary {shapes.misc}
\usepackage[outline]{contour} 
\usepackage{xcolor}
\usepackage{pgfplots}
\pgfplotsset{compat=1.14}

\colorlet{veccol}{green!50!black}
\colorlet{projcol}{blue!70!black}
\colorlet{myblue}{blue!80!black}
\colorlet{myred}{red!90!black}
\colorlet{mydarkblue}{blue!50!black}
\tikzset{>=latex} 
\tikzstyle{proj}=[projcol!80,line width=0.08] 
\tikzstyle{area}=[draw=veccol,fill=veccol!80,fill opacity=0.6]
\tikzstyle{vector}=[-stealth,myblue,thick,line cap=round]
\tikzstyle{unit vector}=[->,veccol,thick,line cap=round]
\tikzstyle{dark unit vector}=[unit vector,veccol!70!black]
\usetikzlibrary{angles,quotes} 
\usetikzlibrary{calc}

\contourlength{1.3pt}

\usepackage{algorithm}

\usepackage{eso-pic}
\usepackage{caption}
\newcommand\AtPageUpperMyright[1]{\AtPageUpperLeft{%
 \put(\LenToUnit{0.17\paperwidth},\LenToUnit{-2cm}){%
     \parbox{0.9\textwidth}{\raggedleft\fontsize{8}{11}\selectfont #1}}%
 }}%
\newcommand{\conf}[1]{%
\AddToShipoutPictureBG*{%
\AtPageUpperMyright{#1}
}
}    

\usepackage[caption=false, font=footnotesize]{subfig}

\begin{document}

\title{\vspace*{1cm} Reinforcement Learning for Gliding Projectile Guidance and Control}

\author{
\IEEEauthorblockN{Cahn Joël}
\IEEEauthorblockA{
\textit{ISAE-SUPAERO}\\
Toulouse, France \\
joel.cahn@student.isae-supaero.fr}
\and
\IEEEauthorblockN{Thomas Antonin}
\IEEEauthorblockA{
\textit{ISAE-SUPAERO}\\
Toulouse, France \\
antonin.thomas@student.isae-supaero.fr}
\and
\IEEEauthorblockN{Pastor Philippe}
\IEEEauthorblockA{
\textit{ISAE-SUPAERO}\\
Toulouse, France \\
philippe.pastor@isae-supaero.fr}
}

\maketitle
\conf{\textit{  V. International Conference on Electrical, Computer and Energy Technologies (ICECET 2025) \\ 
3-6 July 2025, Paris-France}}

\begin{abstract}
This paper presents the development of a control law, which is intended to be implemented on an optical guided glider. This guiding law follows an innovative approach, the reinforcement learning. This control law is used to make navigation more flexible and autonomous in a dynamic environment. The final objective is to track a target detected with the camera and then guide the glider to this point with high precision. Already applied on quad-copter drones, we wish by this study to demonstrate the applicability of reinforcement learning for fixed-wing aircraft on all of its axis.
\end{abstract}


\section{Introduction}
Control of unmanned aircraft that lack propulsion such as gliders presents significant challenges due to their reliance on environmental factors like wind or turbulence. For this kind of aircraft, most traditional control methods often struggle to adapt to the nonlinear and dynamic nature of atmospheric conditions. Upon those traditional methods, we can cite Proportional-Integral-Derivative (PID) controllers or Linear Quadratic Regulators (LQR).
Reinforcement learning, offers a powerful alternative. It is a subfield of machine learning that allows systems to learn optimal control policies by engaging with their environment through simulations.\\
Recent attempts have been made to use RL for UAV control with good success.
It has been widely used in quadcopter control and guidance with great success \cite{song2021autonomous} , sometime even outperforming humans operators \cite{Kaufmann2023}.\\
Reinforcement learning has also been used in fixed wing control \cite{RLReview}, although only few papers conducts real field experiences to test the developed control law. A remarkable attempt \cite{Bohn2023} showed that RL could be as efficient as some state-of-art classical approach such as ArduPlane control laws.
The aim of this studies was to control the attitude of the UAV. However, it was not meant control the aircraft in a higher level such as waypoints navigation or more elaborate objectives.\\
Reinforcement learning has also been proven to be effective in missile guidance. \cite{MissileApproach} showed that a trained agent could learn a policy that would perform as well as state-of-arts approach for missile such as proportional navigation guidance.   

In this paper we will develop a complete framework to apply reinforcement learning algorithms in order to train an agent to control a gliding projectile. The projectile will have to glide as precisely as possible to a designated target.  We will only use an Inertial Measurement Unit (IMU) sensor and a strap-down camera seeker to feed the agent. We also wish to keep the cost of the overall system as low as possible, meaning that we plan to use off the shelf, consumer-grade components.

\section{Modeling the glider}

\subsection{Description of the Glider}

The glider has been modeled as a light fixed-wing UAV. Our prototype has a wingspan of 50 cm and is 60 cm long. As we wanted to be able to rapidly build and test our prototypes, we decided to use 3D printers for the manufacture of the different parts with plastic (PETG). The final weight of the glider is around 300g once the systems and the battery is included.
Our model does not include ailerons on its wing. However, we control pitch and roll with the same moving surfaces, in our case, the horizontal stabilizer which entirely move around an axis situated at 25\% of the horizontal stabilizer's chord. This allows us to have an easy system with a great amplitude.
In order to increase the yaw and pitch we also included 4 stabilizers placed with a 45° angle from each horizontal tail.
Finally, to guide our glider, we placed a camera in the nose of the aircraft.

\subsection{Coordinate System Definition}

\captionsetup{justification=raggedright, labelsep=period}

\tdplotsetmaincoords{59}{120}

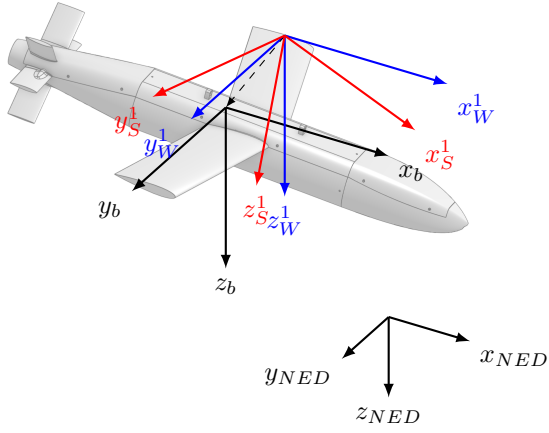
\begin{figure}
    \centering
    \begin{tikzpicture}[scale=2.5,tdplot_main_coords]
      \def\rvec{.8}
      \def\thetavec{30}
      \def\phivec{60}
    
      \node[opacity=0.5] at (0,0,0) {\includegraphics[width=7cm]{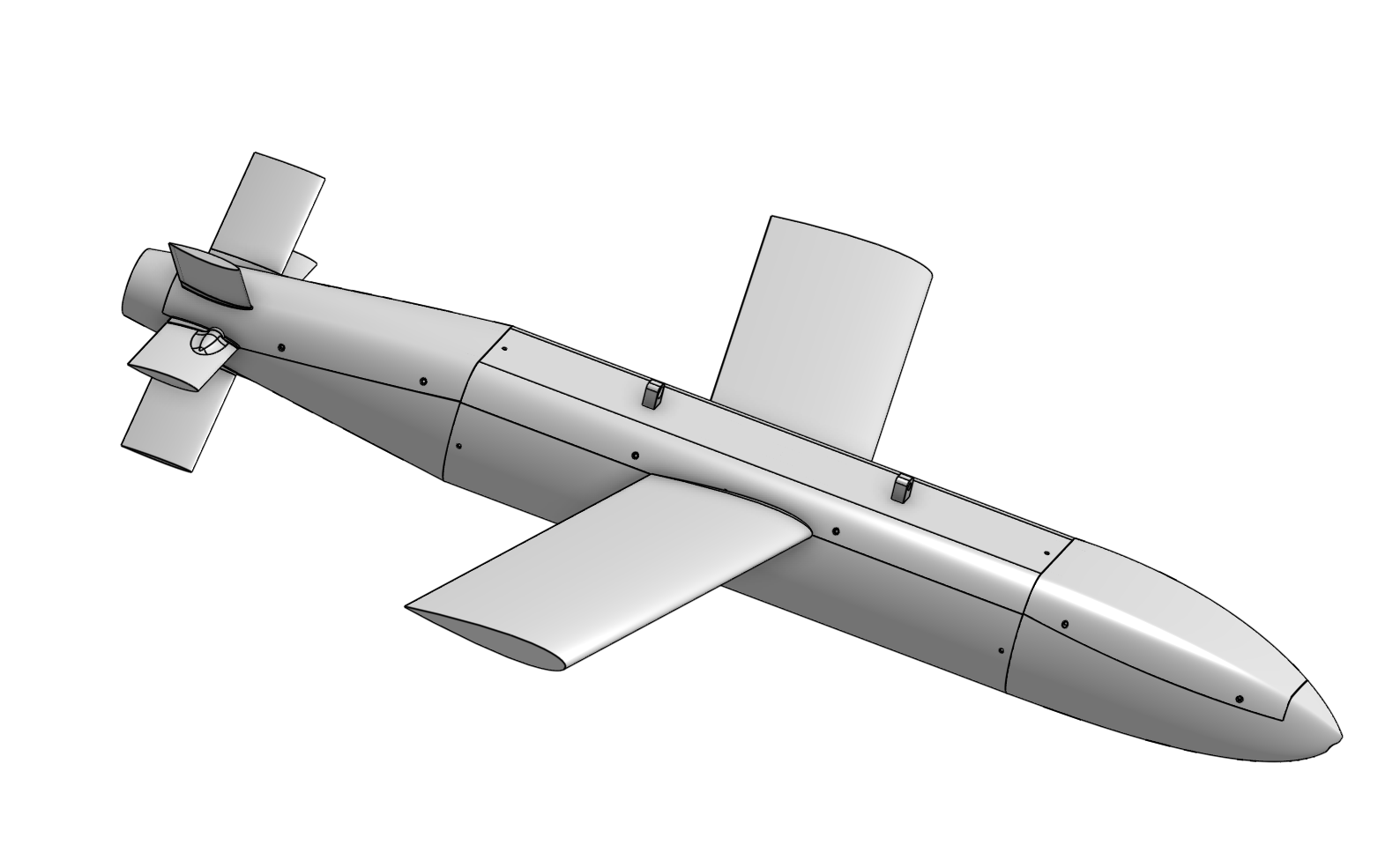}};
      \coordinate (O) at (0,0,0);
      \draw[thick,->] (0,0,0) -- (1.,0,0) node[anchor=north east]{$y_b$};
      \draw[thick,->] (0,0,0) -- (0,1.,0) node[anchor=north west]{$x_b$};
      \draw[thick,->] (0,0,0) -- (0,0,-1.) node[anchor=north]{$z_b$}; 

        \coordinate (SOL) at (0,1,-1);
      \draw[thick,->] (SOL) -- ($(SOL) + (0.5,0,0)$) node[anchor=north east]{$y_{NED}$};
      \draw[thick,->] (SOL) -- ($(SOL) + (0,0.5,0)$) node[anchor=north west]{$x_{NED}$};
      \draw[thick,->] (SOL) -- ($(SOL) + (0,0,-0.5)$) node[anchor=north]{$z_{NED}$};

      \coordinate (A) at (-0.8,-0.1,0);
      \draw[blue,thick,->] (A) -- ($(A) + (1,0,0)$) node[anchor=north east]{$y^1_W$};
      \draw[blue,thick,->] (A) -- ($(A) + (0,1,0)$) node(bob)[anchor=north west]{$x^1_W$};
      \draw[blue,thick,->] (A) -- ($(A) + (0,0,-1)$) node[anchor=north]{$z_W^1$};
    
      \draw[dashed,->] (A) -- (O) ;
      \begin{scope}[rotate around x=-10]
        \begin{scope}[rotate around z=-15]
            \draw[red,thick,->] (A) -- ($(A) + (1,0,0)$) node[anchor=north east]{$y^1_S$};
            \draw[red,thick,->] (A) -- ($(A) + (0,1,0)$) node(xs)[anchor=north west]{$x^1_S$};
            \draw[red,thick,->] (A) -- ($(A) + (0,0,-1)$) node[anchor=north]{$z_S^1$};
        \end{scope}
      \end{scope}
    \end{tikzpicture}
    
    \caption{Coordinate system used in the simulation}

    \label{fig:coordinate-system}
\end{figure}

The attitude is represented using Euler angle $\mathbf{\Theta} = [\phi\ \theta\  \psi]^T$ where $\phi, \theta, \phi$ are the roll, pitch and yaw angle. The position of the center of mass of the glider in the NED frame is noted $\mathbf{p} = [x\ y\ z]^T$.\\
The body frame is fixed relative to the glider, and its origin is located at the center of mass of the glider.\\
The glider is decomposed into lifting surfaces. Each lifting surface has two coordinate systems attached to its neutral point.
The first one called the wing coordinate system $W^i$ is rigidly attached to the wing. The $\mathbf{x_W^i}$ vector is aligned with the chord of the lifting surface, pointing in direction of the leading edge. The $\mathbf{z^i_W}$ vector is normal to the lifting surface, pointing down. The $\mathbf{y^i_W}$ follows the right hand rule, so that $\mathbf{x^i_W} \wedge \mathbf{y^i_W} = \mathbf{z^i_W}$. The transformation from the body coordinate system to $W^i$ is done using the $\mathbf{B2W^i}$ DCM matrix and a translation of $\mathbf{p_W^i}$ from the center of gravity of the glider. This formulation allows us to easily place a lifting surface in all attitudes and locations relative to the center of gravity.\\ 
The second coordinate system is called the wind coordinate system $S^i$. The center of this coordinate system is also located at the neutral point of the lifting surface. In this coordinate system, the $\mathbf{x^i_S}$ is aligned with the airflow that the lifting surface is getting, pointing against the stream. The $\mathbf{y_S^i}$ vector is tangent to the plane $(\mathbf{x_W^i},\mathbf{y_W^i})$. The $\mathbf{z_S^i}$ follows the right-hand rule so that $\mathbf{x^i_S} \wedge \mathbf{y^i_S} = \mathbf{z^i_S}$. The transformation from the $W^i$ to $S^i$ is done using the $\mathbf{W2S^i}$ DCM matrix. This coordinate system is rotated of $\alpha^i_S$ and $\beta^i_S$ respectively around the $\mathbf{y^i_W}$ and $\mathbf{z^i_W}$ vectors. The $\alpha^i_S$ angle is called the local incidence and the $\beta^i_S$ angle is called the sideslip angle. All the coordinate systems introduced in this section are represented by their trihedral in the Fig. \ref{fig:coordinate-system}.
\subsection{6 DOF Equations}
The glider is considered to be a rigid body, with a mass $m$ and an inertia tensor $I$.\\
The speed of the glider can be expressed in the body frame as $\mathbf{v}^b = [u\ v\ w]^T$. The same can be done for the rotational speed denoted $\boldsymbol{\omega}^b = [p\ q\ r]^T$.
It can be shown \cite{ACS} that the flat earth 6 DOF equations are:
\begin{gather}
    \mathbf{\dot{\Theta}} = H(\Theta)\ \omega\\
    \mathbf{\dot{p}} = B\ \mathbf{v}^b\\
    \mathbf{\dot{v}}^b = \frac{1}{m}\ \mathbf{F}^{b}_{Total} + B\ \mathbf{g} - \Omega \mathbf{v}^b \\
    \boldsymbol{\dot{\omega}} = I^{-1}\ [\mathbf{M}^{b}_{Total} - \Omega I\boldsymbol{\omega}]
\end{gather}
with $B$ being the DCM matrix to transform body coordinate to NED, $H$ being the matrix that transforms body rotation rate to Euler rotation rate and $\Omega$ being the skew symmetric matrix of the angular rate in the body frame. $\mathbf{F}^{b}_{Total}$ and $\mathbf{M}^{b}_{Total}$ are respectively the sum of all the efforts and moments expressed in the body frame applied at the center of mass of the solid.

\subsection{Aerodynamic Forces and Moments}
The aero force and moments are computed for each lifting surface. The glider is broken down into six lifting surfaces: the left and right main wing, the left and right elevator, one vertical stabilizer and the body of the glider. For each surface, the local speed is computed using \ref{eq:local_speed}
\begin{equation} \label{eq:local_speed}
    \mathbf{v}^S = \mathbf{v}^b + \boldsymbol{\omega}^b \wedge \boldsymbol{p}_{W}
\end{equation} with $\boldsymbol{p}_{W}$ the position of the lifting surface in the body frame. The local incidence $\alpha_S$ and sideslip $\beta_S$ can be found using $\mathbf{v}^S$
\begin{gather}
    \alpha_S = tan^{-1}(w_S/u_S) \\
    \beta_S = sin^{-1}(v_S/u_S)
\end{gather}
The lift coefficient of the surface is then computed using the 3D lift coefficient equation.\\
\begin{gather} \label{eq:3D_lift_curve}
    C_L = C_{L0} + C_{L\alpha}\alpha\\ 
    C_{L\alpha} = \frac{\pi AR}{1+\sqrt{1+{(\frac{AR}{2})}^2}}
\end{gather}
where $AR$ is the aspect wing ratio of the considered wing.
This coefficient is only valid under the stall incidence. 
In our simulation, we will only take into account the zero-lift drag, $C_{D0}$ and the induced drag $C_{Di}$\\
\begin{gather}
    C_{D} = C_{D0} + C_{Di}\\
    C_{Di} = \frac{C_L^2}{\pi eAR}
\end{gather}
with $e$ the Oswald coefficient of the considered surface. For the sake of simplicity we will take it at 0.75 for each lifting surface.
Then, we compute the wing pitching moment coefficient $C_M$. This coefficient is computed at the aerodynamic center of the wing. 
\begin{equation}
    C_M = C_{M0}
\end{equation}
Since we will be using symmetric airfoils for each wing of the glider, we can say that $C_M=0$ and $C_{L0}=0$.\\
The only aerodynamic coefficient that is taken into account in our simulation for the fuselage of the glider is the drag coefficient $C_{D_{FUS}}$.
\begin{gather}
    C_{D_{FUS}} = FF_W\cdot Cf\cdot \frac{S_{Wet}}{S_{Ref}} 
\end{gather}
with 
\begin{itemize}
    \item $FF_W$ is the form factor of the fuselage
    \item $C_f$ is the skin friction coefficient on the fuselage
    \item $S_{wet}$ is the wet area of the fuselage
    \item $S_{ref}$ is the reference area of the fuselage
\end{itemize}.

The actual efforts and moment for each of the lifting surfaces in the wind frame is computed as:
\begin{gather}
    L = \frac{1}{2}\rho \mathbf{v_S}^2 S C_L\\
    D = \frac{1}{2}\rho \mathbf{v_S}^2 S C_D\\
    M = \frac{1}{2}\rho \mathbf{v_S}^2 cS C_M\\
\end{gather}
with $c$ being the chord of the considered lifting surface.\\
To build up the aero forces of the entire glider, we add up each surface force separately.
\begin{enumerate}
    \item Compute the $\mathbf{B2W^i}$ DCM matrix
    \item Compute $\alpha_S^i,\beta_S^i,V_s^i$
    \item Compute aero forces in the wind frame
    \item Compute the $\mathbf{W2S^i}$ DCM matrix
    \item Transform the effort from the wind frame to the body frame using $\mathbf{B2W^i}$ and $\mathbf{W2S^i}$
    \item Transport aero forces into moment at the center of gravity
    \item Add the force and the transported moment to $\mathbf{F}^{b}_{Total}$ and $\mathbf{M}^{b}_{Total}$
\end{enumerate}

We have implemented a Dryden turbulence model to our simulation. It is based on \cite{abichandani2020wind}  

When the computation of the forces and moments is completed, we integrate the 6DOF equation over a timestep using the Runge-Kutta 4 integration scheme.

This parametric formulation of the glider allows us to quickly change the geometry and adapt the aerodynamic forces and moments automatically. We can also easly compute the forces and moments caused by the two elevon.\\
However, this formulation also has some drawbacks such as the lake of interaction between the different elements of the glider. To keep the simulation simple, we also have not accounted for any side forces and moments other than $L,D,M$.\\
This method provide us with a first approximation of a UAV dynamic model. In some future work, we could implement a more classic approach to get aerodynamics coefficient such as real test identification and/or CFD results to feed our simulation.

\section{Modeling the camera}

\subsection{Objective}

To model the strapdown camera seeker of the gliding projectile, a pinhole camera model has been chosen. It is the simplest camera model that describes the mathematical relationship of the projection of points from 3D space to image space. It is a first good approximation of the seeker in the gliding bomb\\
It is shown that the camera projection could be written in the form:
\begin{equation}
    \mathbf{x} = \mathbf{PX}
\end{equation}
with $\mathbf{x}$ being the coordinate of a point in the image plane in homogeneous coordinates, $\mathbf{X}$ the coordinate of the point in the 3D world in homogeneous coordinate and $\mathbf{P}$ being the projection matrix.
The $\mathbf{P}$ is constructed using intrinsic parameters of the camera, $f$ the focal distance of the camera and $p_x,p_y$ the center of the coordinate system in the image plan and extrinsic parameters that describes the rotation and the translation of the camera in the 3D world space
\begin{equation}
        \mathbf{P} =  
    \left[ {\begin{array}{ccc}
    f & 0 & 0 \\
    p_x & f & 0 \\
    p_y & 0 & 1 \\
  \end{array} } \right] \cdot
  \left[ {\begin{array}{c|c}
    \mathbf{B}  & \mathbf{B} \cdot \mathbf{p}\\
  \end{array} } \right] 
\end{equation}
With this formulation we an extract the position of an object in the screen space by converting the resulting homogeneous coordinates to screen coordinates.\\
We decided not to include a lens distortion model in our camera system, as the distortion could be easily corrected beforehand through a preliminary calibration and preprocessing step.

\usetikzlibrary{positioning}
\section{Reinforcement learning}

Reinforcement learning is a branch of artificial intelligence. In this realm, an agent learns to make decisions by interacting with an environment. The goal of the agent is to maximize a cost function.\\
During the training process, the agent will receive an observation or a state $\mathbf{S_t}$ of the environment. It will then select an action $\mathbf{A_t}$ based only on this observation. This action will lead to a change in the environment. The agent receives the new observation $\mathbf{O_{t+1}}$ and a reward fraction $\mathbf{R_{t+1}}$.\\
Traditionally, RL algorithms tend to maximize the expected discounted cumulative reward
\begin{equation}
    R = \sum_{i=0}^{N} \gamma^i R_i 
\end{equation}
with $N$ final step or the evaluation and $0<\gamma<1$ the discount rate. This parameter will define how "far-sighted" the agent will be. When $\gamma \xrightarrow[]{} 1$ the agent will take each reward into account equally throughout the episode. If $\gamma \xrightarrow[]{} 0$ the agent will tend to prefer immediate rewards over long-term rewards.\\
At the beginning of the training process, the agent has no prior knowledge of the environment. The agent will learn over time to pick the right action based on an observation to maximize the reward.
When the agent is properly trained, it can run on basic hardware without the need of a lot of computational power. It makes it a good candidate for embedded systems. Recent studies have shown that in some cases, an agent trained using reinforcement learning algorithms can outperform humans and other classical controllers \cite{Kaufmann2023}.\\
\subsection{Observation Space}
The observation will define what the perception of the environment is by the agent. In the realm of reinforcement learning, an observation is defined as a normalized vector $\mathbf{O_{t} \in \mathcal{O}}$ with  
\begin{gather}
    \mathcal{O} = \left\{ \mathbf{o} \in \mathbb{R}^n \mid -1 \leq o_i \leq 1, \quad \forall i \in \{1, 2, \dots, n\} \right\}   
\end{gather}
As shown in multiple studies,\cite{Bohn_2019} \cite{RLReview}, a reduced set of observations will lead to better results in terms of convergence time and quality of the trained agent. We have to guess which observation is relevant for the agent to achieve the desired task.\\
For our application, we decide to limit our observation to 6 components:
\begin{gather}
    \mathbf{O_t}= [\phi^{*} ,\theta^{*}, p^{*},q^{*},u^{*}_{cam},v^{*}_{cam}]
\end{gather}
Each component is normalized between -1 and 1 according to the good practice announced by StableBaseline.

\subsection{Action Space}
The action space $\mathcal{A}(s)$ defines what "levers" the agent has on the environment. In our case, it is the symmetrical $\delta_{el}^*$ and asymmetrical $\delta_{ail}^*$ deflection of the "elevons" at the back of the glider. The agent output is normalized between -1 and 1. It is then de-normalized and transformed into an angle to feed the 6DDL simulation.

\subsection{Reward Function}
According to \cite{SILVER2021103535} a reward function can fully define an objective. Our goal is to glide to a target using only a strap-down camera seeker and gyro-accelerometric measurement.\\
The reward function is defined as:
\begin{equation}
    R_{camera} = \omega_1\sqrt{{(u_{target}-u_{center})^2 + (v_{target}-v_{center})^2}}
\end{equation}
\begin{equation}
    R_{actuator} = \omega_2 \delta_{el}^2 + \omega_3 \delta_{ail}^2
\end{equation}
\begin{equation}
    R_t = -(R_{camera} + R_{actuator})
\end{equation}
    
With this formulation of the problem, we train the agent to follow the "LOS" (Line-Of-Sight) guidance. This guidance is not optimal against moving targets or when wind speed is not negligible compared to the glider speed. However, it is efficient against non-moving targets and in calm wind situations \cite{Buc1988} .\\
We added a negative reward directly connected to the actuators position. This term is meant to reduce the actuator deflection and constrain the solution. 

\subsection{Reinforcement Learning Algorithm}

There are many different reinforcement learning algorithms, each with their own specificities. PPO \cite{schulman2017proximalpolicyoptimizationalgorithms} has been proven to be very effective in tasks involving fixed-wing aircraft control \cite{RLReview}. We chose to focus our experimentation on PPO as it is widely and successfully used in aircraft control. We did not change the default hyperparameters provided by the implementation of the PPO algorithm provided by StableBaseline3.

\subsection{Training Environment}

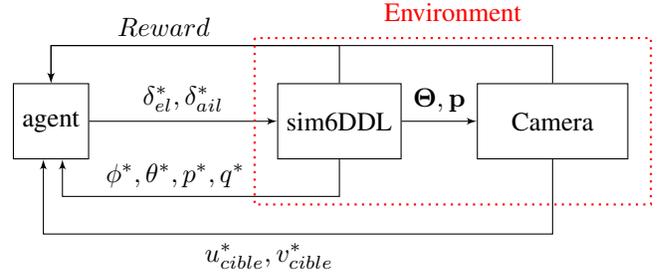
\begin{figure}
        \begin{tikzpicture}[auto, node distance=2cm,>=latex']
            \node [draw, minimum width=1cm, minimum height=1cm] (agent) {agent};
            \node [draw, right=2.5cm of agent,minimum width=1.5cm, minimum height=1cm] (sim6DDL) {sim6DDL};
            \node [draw, right=1cm of sim6DDL, minimum width=2cm, minimum height=1cm] (camera) {Camera};
            \draw[->] ([xshift=0] sim6DDL.north) |- ++(0,0.5) -| node[above,xshift=1.5cm] {$Reward$} (agent.north);
            \draw[->] ([xshift=0mm] camera.north) |- ++(0,0.5) -| (agent.north);
            \draw[->] (sim6DDL) -- node[above,xshift=0.0cm] {$\mathbf{{\Theta,p}}$} (camera);
            \draw[->] (sim6DDL) |- ++(0,-1) -| node[above,xshift=1.5cm] {$\phi^* , \theta^*,p^*,q^*$} ([xshift=3mm] agent);
            \draw[->] (camera) |- ++(0,-1.5) -| node[below, xshift=3cm] {$u_{cible}^* , v_{cible}^*$} ([xshift=-3mm] agent);
            \draw[->] (agent) -- node[above] {$\delta_{el}^*,\delta_{ail}^*$} (sim6DDL);
            \draw[red,thick,dotted] ($(sim6DDL.north west)+(-0.3,0.6)$)  rectangle node[above,yshift=12mm ]{Environment} ($(camera.south east)+(0.3,-0.6)$) ;
        \end{tikzpicture}
    \caption{Training framework organisation}
    \label{fig:RL-framework}
    \end{figure}

We have set up an environment for training our agent. It is organized as shown in the Fig. \ref{fig:RL-framework}. We have created a custom Gymnasium \cite{towers2024gymnasiumstandardinterfacereinforcement} environment with our implementation of the 6DDL simulation loop and the camera. \\
We are using the library StableBaseline3 \cite{stable-baselines3} for its implementation of a reinforcement learning algorithm.
The simulation loop is set at 100 Hz and the agent is refreshed at every update of the environment. The observation vector is also noise-free and without lag.

We initialize the position of the glider and the target according to Table \ref{tab1}. The angle of the cone $\theta_{init}^{NED} = atan2(x_{init},y_{init})$ represents the angle between the initial position of the glider and a vector facing the North. We randomize the initial position of the glider so that the agent can learn as much as possible about its environment. We must ensure that the task described by the reward function is achievable for each initialization case. In our case, we give our glider enough potential and kinetic energy to let it maneuver towards the target. We also have to ensure that the target is within the field of view of the camera at the beginning of each episode.
\captionsetup{justification=centering, labelsep=period}
\begin{table}[htbp]
\begin{center}
    \caption{Episode initial conditions}

    \begin{tabular}{|c|c|c|}
        \hline
        \textbf{Variable} & \textbf{Mean value}& \textbf{Random range}\\ 
        \hline 
        \hline
        $r_{init}$ & 150m & $\pm 50$ m \\
        $\theta_{init}^{NED}$ & $0\deg$ & $\pm 45\deg$ \\
        $z_{init}$  & $ r_{init}/2$ & $\pm 20$ m \\
        $\phi_{initial}$ & $0\deg$ & $0\deg$ \\ 
        $\psi_{initial}$ & $0\deg$ & $0\deg$ \\ 
        $x_{target}$ & 0 m & 0 m \\
        $y_{target}$ & 0 m & 0 m \\
        $z_{target}$  & 0 m & 0 m \\
        \hline
        \end{tabular}
    \label{tab1}
\end{center}
\end{table}
\\
\captionsetup{justification=raggedright,singlelinecheck=false,labelsep=period}

At the beginning of each episode the glider is trimmed so:
\begin{gather}
    \delta_{el}^{trimed} = 0    
\end{gather}
This automatically set $\mathbf{v^b}_{initial}$ and $\theta_{initial}$. For the moment we have not included any disturbance in our environment, there is no turbulence nor wind.

\tikzstyle{block} = [draw, rectangle, 
    minimum height=3em, minimum width=6em]
\tikzstyle{sum} = [draw, circle, node distance=1cm]
\tikzstyle{input} = [coordinate]
\tikzstyle{output} = [coordinate]
\tikzstyle{pinstyle} = [pin edge={to-,thin,black}]

\section{Classic controller}

To our knowledge, there is not another paper on exactly the same problematic with our result metrics. To get a comparison with our results, we want to use another type of control. That is the reason why we decided to use a more standard controller in the same environment to see what are the differences with reinforcement learning. We expect to see changes in the action, the precision and the dispersion.\\
The "classical" solution consist of two distinct controller, one for the longitudinal control and one for the lateral control. \\
In order to decouple the camera output into longitudinal and lateral channel for our controllers, we de-rotate the camera around the roll axis:
\begin{equation}
    [u^*_{stab},v^*_{stab}]^T =  
    \left[ {\begin{array}{cc}
    cos(\phi) & -sin(\phi) \\
    sin(\phi) & cos(\phi) \\
  \end{array} } \right] \cdot
  \left[ {\begin{array}{c}
    u^*_{cible}\\
    v^*_{cible} \\
  \end{array} } \right] 
\end{equation}

The longitudinal control is a simple PID controller that takes the $v^*_{stab}$ channel of the seeker and transform it to a symmetrical command deflection $\delta_{el}^*$.\\
The lateral control consist of two cascading PID controllers. The first one is a "heading" controller. The input of this controller is the $u^*_{stab}$ channel of the seeker and it output a roll command. The second PID is linked to the first one, and function as a roll controller. It output the asymmetrical deflection of the elevon $\delta_{ail}^*$.

\section{Results and discussion}

\subsection{Convergence}
The first thing to check during the training of an agent is the convergence of the solution. In fact, the episodic reward has to consistently improve over the training duration. We also favor a quick convergence as it means that we will not need a lot of computational power to train our agent in a reasonable amount of time.\\

\begin{figure}[h]
    \begin{center}
        \scalebox{0.6}{\input{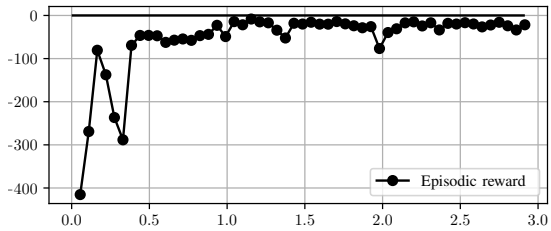}}
    \end{center}
    \caption{Episodic reward over training}
    \label{fig:conv}
\end{figure}

Fig. \ref{fig:conv} shows that with our implementation of the simulation and reward function, we reach an asymptotic episodic reward in around 25 minutes. We achieved this result in a little over 1 million time steps at 100Hz. We are using a laptop equipped with an AMD Ryzen 7 5800 CPU, an NVidia RTX3060 GPU and 32Go of RAM . Training is carried out in 55 environments running in parallel using the CUDA framework.

\subsection{Trajectory Analysis and Comparison}

\begin{figure}[h]
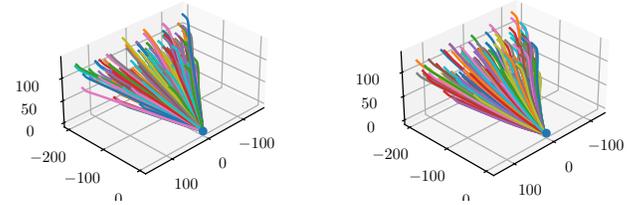

    \subfloat[No wind\label{1a}]{
    \begin{adjustbox}{clip,trim=1cm 0cm 2cm 0cm}
        \scalebox{0.6}{\input{Graph/3D.pgf}}
    \end{adjustbox}
    }
    \subfloat[3 $m.s^{-1}$ east wind\label{1b}]{
        \begin{adjustbox}{clip,trim=1cm 0cm 1cm 0cm}
            \scalebox{0.6}{\input{Graph/3D_wind.pgf}}
        \end{adjustbox}
    }
    \caption{Trajectory of the RL agent}
    \label{fig:3D_traj}
\end{figure}

As we can see in the Fig. \ref{fig:3D_traj}, the agent behave as expected. After the initialization, it directly take the direction of the target. When the wind is not null, the trajectory is shifted in the wind direction. In fact, the agent tries to align the nose of the glider with the target on camera at every timestep. It follows a specific curve, called the "pursuit curve". In the context of an interception engagement, this trajectory is not optimal, leading to lot of dispersion and high actuator action at the end of an episode.

\begin{figure}[h]
    \begin{center}
        \scalebox{0.6}{\input{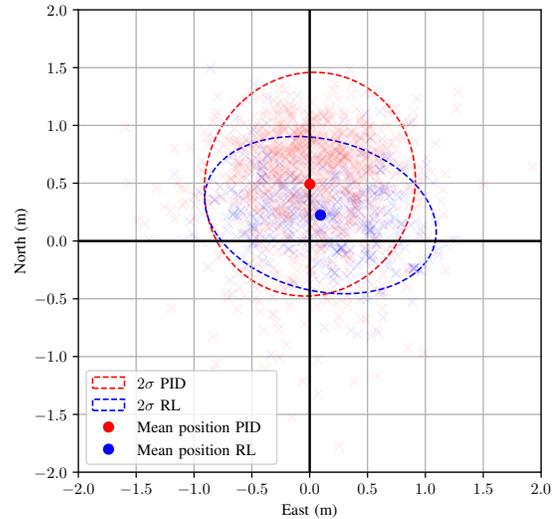}}
        \caption{Circular Error Probable (CEP) no wind}
        \label{fig:CEP_no_wind}
    \end{center}
\end{figure}

\begin{figure}[h]
    \begin{center}
        \scalebox{0.6}{\input{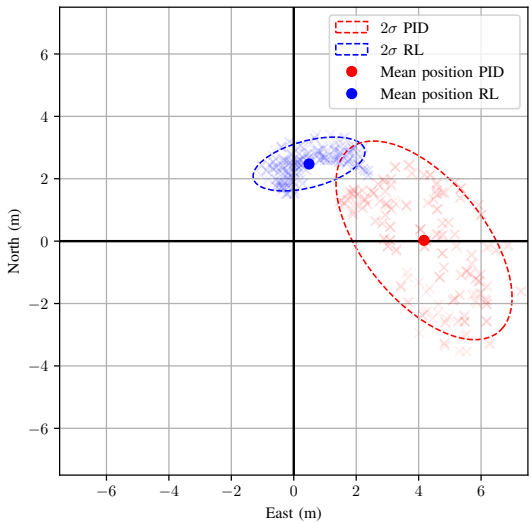}}
        \caption{CEP wind 3 $m.s^{-1}$}
        \label{fig:CEP_wind}
    \end{center}
\end{figure}
\captionsetup{justification=centering, labelsep=period}

\begin{table}[htpb]
\caption{Mean miss distance (MMD) and deviation}
\centering
\begin{tabular}{|c|c|c|}
\hline
\multicolumn{3}{|c|}{\textbf{RL}}\\
\hline
& \multicolumn{1}{|c|}{\textbf{No wind}} & \multicolumn{1}{|c|}{\textbf{East wind $3m.s^{-1}$}} \\
\cline{1-3}
MMD (m)& 0.243 & 2.52\\
${2\sigma_{x,y}}$ (m)& 1.00;0.680 & 1.792;0.864 \\
\hline
\hline
\multicolumn{3}{|c|}{\textbf{PID}}\\
\hline
MMD (m)& 0.492 & 4.173\\
${2\sigma_{x,y}}$ (m)& 0.913;0.968 & 2.820;3.185 \\
\hline
\end{tabular}
\label{tab:CEP_results}
\end{table}

\captionsetup{justification=raggedright,singlelinecheck=false,labelsep=period}

In the case where the wind is not null, there is more dispersion and the mean miss distance is greater than the case without wind. This result is expected since we are in a case of a "LOS" guidance that is not effective where the target is moving or when there is wind.\\
We can also see in the Table \ref{tab:CEP_results} that the RL agent is a little more accurate than the PID law when no wind is present. When wind is present, we can see that the RL agent outperform the PID law in both accuracy and dispersion. Those result can be visually seen in the Fig. \ref{fig:CEP_no_wind} and Fig. \ref{fig:CEP_wind}.

\section{Conclusion and future work}

In this paper we have developed a six degree of freedom simulation environment. We are using the Runge-Kutta 4 integration scheme to integrate the simulation. The aerodynamic model of the glider is constructed procedurally. The glider is decomposed into lifting surfaces, the local airspeed is computed for each lifting surface to determine the forces produced by the surface. Forces are then reported on the center of gravity of the glider. We developed a camera model to simulate the camera seeker of the glider in a virtual environment. A new Gymnasium environment has been created to integrate our modeled glider into a reinforcement learning framework compatible with the StableBaseline3 library. We then crafted a reward function adapted to our need. To design this reward function, we simplified the initial problem into a "LOS" guidance scheme. A training has been carried out using the PPO algorithm and we obtained a good convergence in a limited amount of time. The trained agent is then evaluated using the Monte-Carlo method to determine the dispersion of the glider impact relative to the target. This result is compared to another method we developed in a earlier work and prove the benefice of the reinforcement learning.\\
We also saw that the developed agent was robust to turbulence, as we trained it and tested it using the Dryden turbulence model. However, it was not robust against constant wind due to the type of guidance law it learned. We could mitigate this issue by making learn another type of guidance such as Proportional Navigation. This guidance law allows the agent to effectively intercept a moving target or be robust constant wind.\\
Another axis of improvement would be to take in account in our simulation the actuator dynamics and observe the agent's performances. We could also add disturbance such as noise in the observation vector to evaluate the agent in a more realistic environment.\\
Finally, we plan to implement the trained agent into a radio controled model to compare our parametric aerodynamic coefficient build-up method and a real model. We also plan to test the trained agent on this real model.

\bibliographystyle{ieeetr}


\vspace{12pt}

\end{document}